\documentclass[10pt,twocolumn,letterpaper]{article}

\usepackage[pagenumbers]{wacv} %

\definecolor{wacvblue}{rgb}{0.21,0.49,0.74}
\usepackage[pagebackref,breaklinks,colorlinks,allcolors=wacvblue]{hyperref}
\usepackage[table]{xcolor}
\definecolor{OursRow}{RGB}{222,235,247}

\usepackage[final]{microtype} %
\usepackage{multirow} 
\usepackage{bbm}

\def\wacvPaperID{2093} %
\def\confName{WACV}
\def\confYear{2027}

\title{BRepA: Graph-Based B-Rep Pretraining for Unified CAD\\ Understanding and Generation}

\author{
Dekai Zhu$^{1,2,3}$, ~
Zhaiyu Chen$^{1,3}$, ~
Stefan Gavranovic$^{2}$, ~
and~ 
Slobodan Ilic$^{1, 2}$\\
\\[-0.5em]
$^1$ Technical University of Munich\quad
$^2$ Siemens AG\quad
$^3$ Munich Center for Machine Learning\\
\small{\texttt{\{firstname.lastname\}@tum.de}} \\
}

\begin{document}
\maketitle
\begin{abstract}

Boundary representation (B-Rep) of CAD models encodes geometric primitives and their relations as a graph structure, enabling efficient shape modeling within CAD software. Learning on B-Reps for downstream tasks, such as segmentation and novel shape generation, requires specialized neural network designs. Moreover, performance improvements and accelerated training can be achieved with effective pretraining strategies, which is particularly important for tasks with limited labeled B-Rep data.
In this paper, we introduce \textbf{BRepA}, a unified self-supervised pretraining framework that explicitly reconstructs both the geometry and topology of B-Rep objects via graph-based modeling. By learning from large-scale unlabeled CAD data only once, BRepA produces representations that generalize across tasks and significantly enhance both discriminative and generative learning.
On segmentation tasks, BRepA demonstrates remarkable data efficiency. For instance, with only 1\% labeled samples, BRepA pretraining improves mIoU by \textbf{32.52\%} compared to models trained from scratch. On generative tasks, aligning diffusion model representations with pretrained BRepA features substantially accelerates convergence, achieving nearly \textbf{2$\times$ faster} training while improving generation quality under the same computational budget, despite the diffusion model being roughly \textbf{20$\times$ larger} than BRepA.
Overall, our results show that geometry- and topology-aware B-Rep pretraining provides a powerful foundation for both understanding and generating CAD structures, serving as a versatile representation backbone for diverse downstream applications.
We will release the code to facilitate future research.

\end{abstract}

\section{Introduction}
\label{sec:intro}

Boundary representation (B-Rep) is a fundamental CAD data format that encodes both the geometric and topological structure of 3D objects~\cite{uvnet, brepnet}.
The geometric information describes the parametric characteristics of surfaces, edges, and vertices, while the topological information defines both inter-type relationships (e.g., which edges bound a face) and intra-type adjacencies (\eg, neighboring faces or edges).
B-Rep is widely used in industrial design and manufacturing workflows for visualization, machining feature management, and engineering analysis.

Despite its importance, training B-Rep models presents unique challenges. Models like UV-Net~\cite{uvnet}, BRepNet~\cite{brepnet} and SB-GCN~\cite{jones2021sb} are examples of neural network architectures operating on B-Reps. Compared to 2D images or 3D data~\cite{zhu2025sealion, Zhu_2025_BMVC, zhu2025spiral}, the amount of publicly available B-Rep data is limited, and labeled datasets are particularly scarce, making B-Rep models prone to overfitting on small datasets~\cite{jones2023self}.
Moreover, B-Rep is a parametric and symbolic representation, which is highly abstract and discrete in the feature space.
This discrete and fragmented data manifold makes it especially challenging to train generative models such as diffusion models~\cite{ho2020denoising}, which typically rely on continuous and smooth spaces.

\begin{figure}[t]
    \begin{center}
    \includegraphics[width=\linewidth]{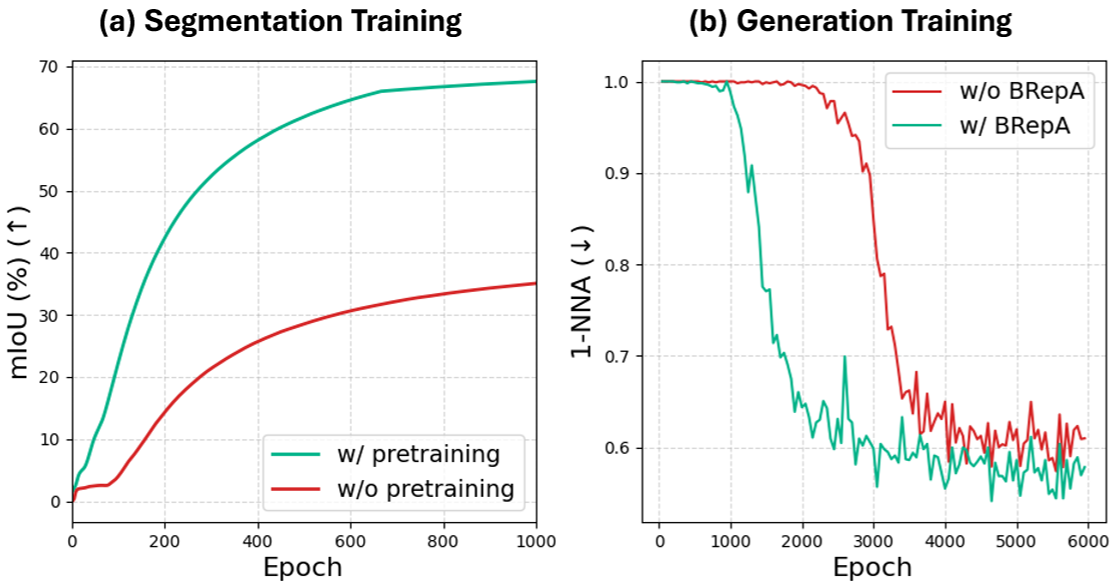}
    \caption{(a) \textbf{Segmentation Training:} With geometry and topology MAE pretraining, the BRepA model equipped with the segmentation head ($MLP_{seg}$) achieves a \textbf{32.52\%} mIoU improvement when labeled samples are rare (1\%). 
    ~(b) \textbf{Generation Training:} Leveraging feature alignment with BRepA, BrepDiff~\cite{brepdiff} trains nearly \textbf{twice as fast} and converges to a better checkpoint, even though BrepDiff (260M) is roughly \textbf{20$\times$ larger} than BRepA (13.5M).
    Both experiments are conducted on the {MFCAD}~\cite{cao2020graph} dataset.
}
    \label{fig:teaser}
    \end{center}
\end{figure}

A promising strategy to address these challenges is self-supervised pretraining, where models are trained on available unlabeled B-Rep data to learn generalizable representations, which can then be fine-tuned or aligned for downstream tasks.
However, research in this direction remains limited.
To our knowledge, only~\citep{jones2023self, qin2025cadgcl, quan2024self} have explored self-supervised pretraining on B-Rep data, and its effectiveness has been evaluated only on discriminative tasks.

In this work, we introduce \textbf{BRepA}, a new B-Rep pretraining model designed to learn both geometric and topological representations.
BRepA naturally models B-Rep data as graphs by using Graph Attention models (GAT)~\cite{velivckovic2017graph} and adopts Transformer-based encoders for both surfaces and edges.
It jointly learns geometry- and topology-aware representations through two complementary pretraining objectives: \textit{Geometry Masked Autoencoding (MAE)} and \textit{Topology MAE}.
Our experiments demonstrate that the learned representations from BRepA benefit both discriminative and generative tasks:
\textbf{(1)} In segmentation tasks, pretrained BRepA consistently outperforms both untrained BRepA and UV-Net, particularly in low-label regimes, as depicted in Figure~\ref{fig:teaser}~(a).
\textbf{(2)} In generative tasks, BRepA enables faster and more effective training of large-scale generative models such as BrepDiff~\cite{brepdiff} through feature alignment, improving both training efficiency and final generation quality, as depicted in Figure~\ref{fig:teaser}~(b), even though these diffusion-based models contain far more parameters than BRepA itself.

In summary, this work makes the following contributions:

\begin{itemize}
    \item \textbf{Novel Architecture}. BRepA employs ViT-based encoders for geometric features of surfaces and edges, and GAT-based encoders for topological features, achieving superior performance compared to CNN- and vanilla-GCN-based UV-Net under similar model sizes.
    \item \textbf{Dual MAE Pretraining}. BRepA jointly learns geometric and topological representations via two complementary reconstruction tasks, both of which enhance downstream performance.
    \item \textbf{Cross-Domain Generalization}. The pretrained BRepA benefits both discriminative and generative tasks, improving segmentation accuracy and accelerating the training of state-of-the-art B-Rep diffusion models.
\end{itemize}
\section{Related Work}
\label{sec:related_work}

\paragraph{Self-Supervised Pretraining.}
Self-supervised learning (SSL) has become a powerful strategy for learning transferable representations without manual annotations~\cite{he2020momentum, chen2020simple, caron2021emerging, hou2022graphmae}.
DINO-v2~\cite{oquab2023dinov2} further demonstrates that large-scale SSL can produce semantically rich features that generalize across downstream tasks.
Recent Representation Alignment (REPA) shows that aligning diffusion model features with DINO-v2 representations accelerates convergence and improves generation quality~\citep{repa}.
SSL on B-Reps has been studied in~\cite{jones2023self, qin2025cadgcl, quan2024self}. 
\cite{qin2025cadgcl, quan2024self} focus on global retrieval, while~\citep{jones2023self} addresses segmentation via geometric reconstruction.
Yet,~\citep{jones2023self} only reconstructs corrupted surface features and does not explicitly capture topological relationships.

\paragraph{B-Rep Segmentation.}
B-Rep segmentation is a fundamental task in CAD understanding, aiming to assign semantic or functional labels to surfaces or faces within B-Rep models~\cite{zhang2018featurenet, angrish2019fabsearch, uvnet, koch2019abc, lee2023brepgat, colligan2022hierarchical, lou2023brep}.
Two widely used datasets for this task are Fusion 360 Segmentation~\cite{willis2020fusion} and MFCAD~\cite{cao2020graph}, which provide annotated B-Rep models of mechanical parts, enabling the development and evaluation of learning-based methods. Early approaches such as SB-GCN employed hierarchical graph convolutional networks to jointly encode geometric and topological information~\cite{jones2021sb}, while UV-Net introduced a hybrid architecture combining convolutional and graph-based encoders to better capture both local surface geometry and global connectivity~\cite{uvnet}.

\paragraph{B-Rep Generation.}
B-Rep generation has become an active direction in neural CAD, aiming to reconstruct B-Reps from geometric inputs or synthesize them from learned distributions~\cite{wu2021deepcad, guo2022complexgen, jayaraman2022solidgen}.
Recent methods largely build on diffusion models~\cite{ho2020denoising, nichol2021improved, guo2025brepgiff, xu2022skexgen}, following either cascade or one-step inference method.
Cascade methods such as BrepGen~\citep{brepgen} progressively generate surface and edge bounding boxes and latent features, which are then decoded into complete B-Rep structures.
In contrast, one-step methods such as BrepDiff~\citep{brepdiff} directly generate surface UV-grid coordinates and visibility masks, followed by post-processing to recover B-Reps.
Other works further explore holistic latent representations and alternative diffusion formulations for B-Rep synthesis~\cite{liu2025hola, li2025dtgbrepgen, li2025stitch}.

\section{Method}

\begin{figure*}[th]
    \centering
    \begin{center}   \includegraphics[width=0.95\linewidth]{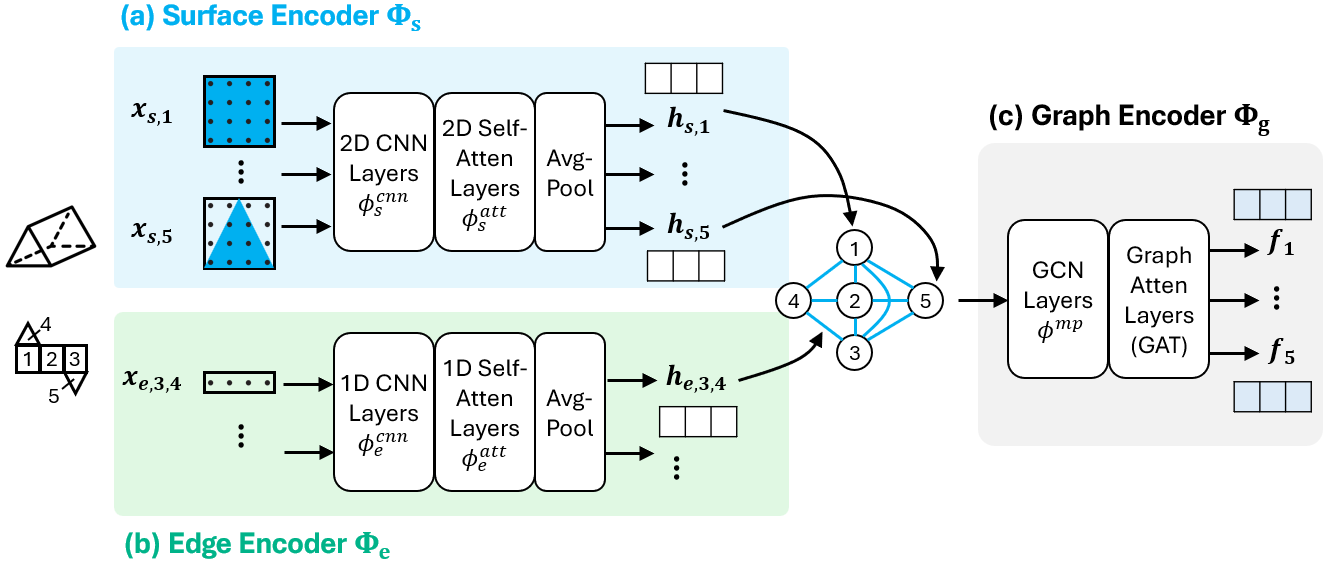}
    \caption{\textbf{Overview of BRepA Architecture.} (a) Surface Encoder $\mathbf{\Phi_s}$ processes 2D surface grids $\mathbf{x}_{s,i}$ using 2D CNN layers with self-attention and average pooling to extract surface embedding $\mathbf{h}_{s,i}$. (b) Edge Encoder $\mathbf{\Phi_e}$ employs 1D CNN layers with self-attention to encode 1D edge grids $\mathbf{x}_{e,i,j}$ and generate edge embedding $\mathbf{h}_{e,i,j}$. (c) Graph Encoder $\mathbf{\Phi_g}$ utilizes message-passing layers of Graph Convolutional Networks (GCN)~\cite{scarselli2008graph} and Graph Attention Layers~\cite{velivckovic2017graph} to capture topological relationships and produce representations $\mathbf{f}_i$.}
    \label{fig:brepa_arch}
    \end{center}
\end{figure*}

In this section, we first provide the preliminaries on B-Rep data representation. 
Next, we introduce the architecture of BRepA, including the surface, edge, and graph encoders, 
as well as its dual masked autoencoding pretraining scheme on both geometric and topological levels. 
Finally, we describe how BRepA is utilized for training both discriminative and generative models.

\subsection{Preliminaries}

Formally, each B-Rep can be expressed as a graph $G(V, E)$ that models both the geometry and topology of a 3D object, where the vertices $V$ correspond to faces and the edges $E$ encode the connectivity between adjacent faces. 
We follow the UV-Net representation~\cite{uvnet} for B-Rep geometry, where the geometry of a curve $C(u)$ is discretized over its parametric domain $[u_{\min}, u_{\max}] \in \mathbb{R}$ into a regular 1D grid with a uniform step size 
$\delta_u = \frac{u_{\max} - u_{\min}}{M - 1}$, where $M$ is the number of sampled points. 
Similarly, the geometry of a surface $S(u, v)$ is discretized over its parametric domain 
$[u_{\min}, u_{\max}] \times [v_{\min}, v_{\max}] \in \mathbb{R}^2$ into a regular 2D grid with step sizes 
$\delta_u = \frac{u_{\max} - u_{\min}}{H - 1}$ and $\delta_v = \frac{v_{\max} - v_{\min}}{W - 1}$, 
where $H$ and $W$ denote the number of samples along each parameter dimension. 
Consequently, each surface feature is represented as $\mathbf{x}_s \in \mathbb{R}^{H \times W \times d_s}$, 
while each edge feature is represented as $\mathbf{x}_e \in \mathbb{R}^{M \times d_e}$,
where $d_s$ and $d_e$ denote the feature dimensions for vertices and edges, respectively.
$\mathbf{x}_s$ contains 3D coordinates, normals, and trimming masks, while $\mathbf{x}_e$ contains 3D coordinates and tangents. Therefore, we have $d_s = 7$ and $d_e = 6$.

\subsection{BRepA Architecture}
The encoding of B-Rep involves processing surface features $\mathbf{x}_s$, edge features $\mathbf{x}_e$, and the graph $G(V, E)$. 
In our BRepA framework, as shown in Figure~\ref{fig:brepa_arch}, we follow the design architecture of ConViT~\cite{d2021convit}, which combines convolutional and transformer-based modules to perform local and global level modeling, respectively. 
This hybrid design leverages the lightweight advantage of CNNs for efficient local feature extraction and the superior global awareness of Vision Transformers~\cite{dosovitskiy2020image}.

\begin{figure*}[t]
    \begin{center}   \includegraphics[width=0.9\linewidth]{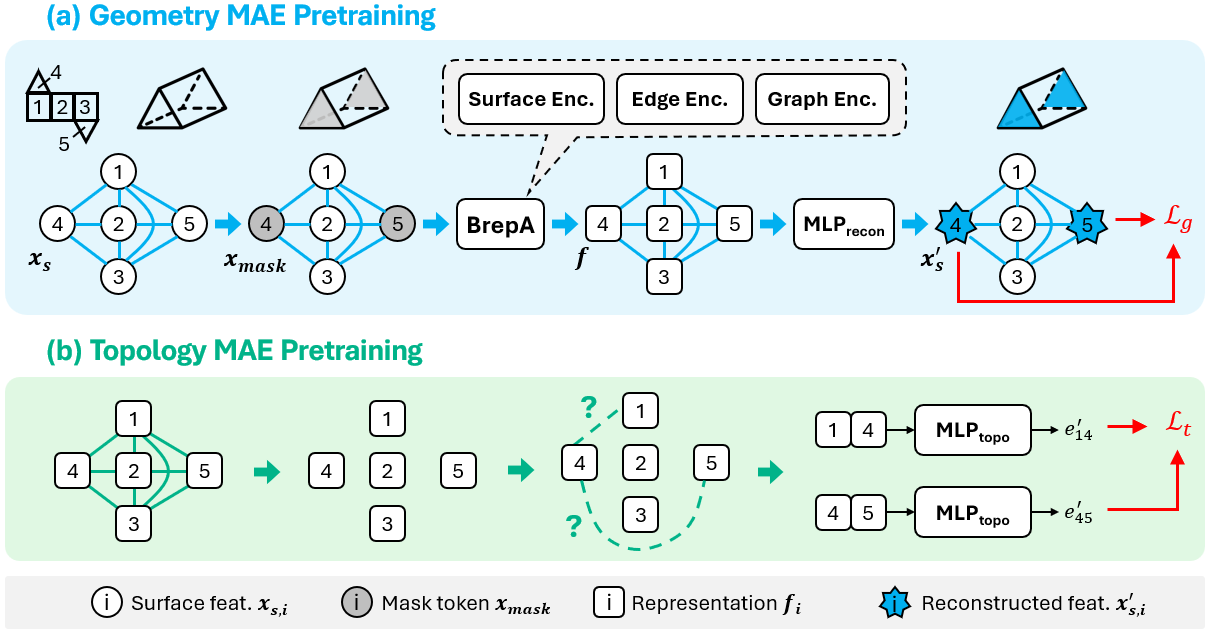}
    \caption{\textbf{Overview of the Dual MAE Pretraining Pipeline in BRepA.}
    (a) The geometry MAE reconstructs masked surface feature~$\mathbf{x}'_{s,i}$ from representation~$\mathbf{f}_{i}$ of node~$i$ encoded by BRepA, which contains surface, edge, and graph encoders.
    (b) The topology MAE predicts pairwise connectivity between surface nodes based on their encoded representations.
    By learning to infer missing edges, it enables BRepA to capture the underlying topological structure and relational dependencies among surfaces.}
    \label{fig:mae}
    \end{center}
\end{figure*}

\paragraph{Surface Encoder $\Phi_s$.}
Given the input surface features $\mathbf{x}_s$, the surface encoder $\Phi_s$ first applies three 2D convolutional layers to capture local geometric patterns and increase the feature dimensionality. 
Subsequently, a stack of self-attention layers is employed across the entire surface to model long-range dependencies. 
Finally, an average pooling layer aggregates the token-level representations into a compact surface embedding $\mathbf{h}_s \in \mathbb{R}^{d_{h_s}}$:
\begin{equation}
\mathbf{h}_s = \mathrm{AvgPool} \big( \phi_s^{\text{att}}(\,\phi_s^{\text{cnn}}(\mathbf{x}_s)\,) \big),
\end{equation}
where $\phi_s^{\text{att}}$ denotes the self-attention layers and $\phi_s^{\text{cnn}}$ denotes the 2D CNN layers.

\paragraph{Edge Encoder $\Phi_e$.}
Similarly, given the input edge features $\mathbf{x}_e$, the edge encoder $\Phi_e$ consists of three 1D convolutional layers followed by three self-attention layers and an average pooling operation, producing the edge embedding $\mathbf{h}_e \in \mathbb{R}^{d_{h_e}}$:
\begin{equation}
\mathbf{h}_e = \mathrm{AvgPool} \big( \phi_e^{\text{att}}(\,\phi_e^{\text{cnn}}(\mathbf{x}_e)\,) \big),
\end{equation}
where $\phi_e^{\text{att}}$ denotes the self-attention layers and $\phi_e^{\text{cnn}}$ denotes the 1D CNN layers.

\paragraph{Graph Encoder $\Phi_g$.}
After encoding the surfaces and edges, the topological graph $G(V, E)$ contains the node features $\mathbf{h}_s$ and edge features $\mathbf{h}_e$. 
BRepA performs message passing between nodes and edges with a hop number of three to propagate information through the topology. 
Finally, the node features are fed into a 3-layer Graph Attention Network (GAT)~\cite{velivckovic2017graph} to obtain the final encoded node representations $\mathbf{f} \in \mathbb{R}^{d_f}$:
\begin{equation}
\mathbf{f} = \mathrm{GAT} \big( \phi^{\text{mp}}(G, \mathbf{h}_s, \mathbf{h}_e) \big),
\end{equation}
where $\phi^{\text{mp}}$ denotes the GCN~\cite{scarselli2008graph} node-edge message passing module.

\subsection{Dual MAE Pretraining}
\label{subsec:dual_mae}
BRepA follows the Masked Autoencoding (MAE) strategy for pretraining.
Unlike prior work~\cite{jones2023self} that reconstructs only geometric features, 
BRepA jointly reconstructs geometry and topology.
This \textit{dual MAE} design, shown in Figure~\ref{fig:mae}, encourages expressive and topology-aware representations.

\paragraph{Geometry MAE Pretraining.}
Given a B-Rep graph $G(V, E)$, BRepA randomly masks a subset $\mathcal{M}$ of surface nodes by replacing their surface features $\mathbf{x}_{s,i}$ with a learnable mask token $\mathbf{x}_{\text{mask}}$.
The corrupted graph is encoded by BRepA to obtain node representations $\mathbf{f}_i$, which are then passed to $\mathrm{MLP}_{\text{recon}}$ to reconstruct the original surface features $\mathbf{x}'_{s,i}$.
The geometry loss $\mathcal{L}_{geo}$ is computed only on the masked nodes.
\begin{align}
\mathbf{x}_{s,i}^{\,m} &=
\begin{cases}
\mathbf{x}_{\text{mask}}, & \text{if } v_i \in \mathcal{M}, \\
\mathbf{x}_{s,i}, & \text{otherwise},
\end{cases} \\[4pt]
\mathbf{f}_i &= \mathrm{BRepA}\big(G, \{\mathbf{x}_{s}^{\,m}\}, \{\mathbf{x}_{e}\}\big), \\[4pt]
\mathbf{x}'_{s,i} &= \mathrm{MLP}_{\text{recon}}(\mathbf{f}_i), \\[4pt]
\mathcal{L}_{geo} &= \frac{1}{|\mathcal{M}|} \sum_{v_i \in \mathcal{M}} 
\mathrm{MSE}\big(\mathbf{x}_{s,i},\, \mathbf{x}'_{s,i}\big).
\end{align}

\paragraph{Topology MAE Pretraining.}
The topological structure of $G(V, E)$ encodes the connectivity relationships among surface nodes. 
To learn topology-aware representations, BRepA first randomly drops edges in the graph $G$ to perturb message passing. It then samples pairs of nodes $(v_i, v_j)$ and predicts their connectivity based on the corresponding encoded node features $\mathbf{f}_i$ and $\mathbf{f}_j$.
For each node pair, their features are concatenated and fed into a topology prediction head $\mathrm{MLP}_{\text{topo}}$ to estimate the probability of an existing edge $e'_{ij} \in [0, 1]$. 
The topology reconstruction loss $\mathcal{L}_{topo}$ is computed only over the sampled node pairs:
\begin{align}
\mathbf{f}_{ij} &= \mathbf{f}_i \oplus \mathbf{f}_j , \\[4pt]
e'_{ij} &= \mathrm{MLP}_{\text{topo}}(\mathbf{f}_{ij}), \\[4pt]
\mathcal{L}_{topo} &= \frac{1}{|\mathcal{S}|} \sum_{(i,j)\in\mathcal{S}}
\mathrm{BE}\big(e_{ij},\, e'_{ij}\big),
\end{align}
where $\mathcal{S}$ denotes the set of sampled node pairs, 
and $\mathrm{BE}(\cdot)$ represents the binary cross-entropy loss between the ground-truth connectivity $e_{ij}$ and the predicted probability $e'_{ij}$.

\paragraph{Overall Pretraining Objective.}
BRepA is pretrained by jointly optimizing the geometric and topological reconstruction objectives. 
The overall loss is defined as
\begin{equation}
\label{eq:overall_mae}
\mathcal{L}_{MAE} = \mathcal{L}_{geo} + \alpha \, \mathcal{L}_{topo},
\end{equation}
where $\alpha$ is a weighting coefficient that balances the contributions of the geometric reconstruction loss $\mathcal{L}_{geo}$ and the topological reconstruction loss $\mathcal{L}_{topo}$.
This dual-objective design enables BRepA to learn representations that are simultaneously geometry-aware and topology-aware.

\begin{figure}[t]
    \begin{center}
    \includegraphics[width=0.9\linewidth]{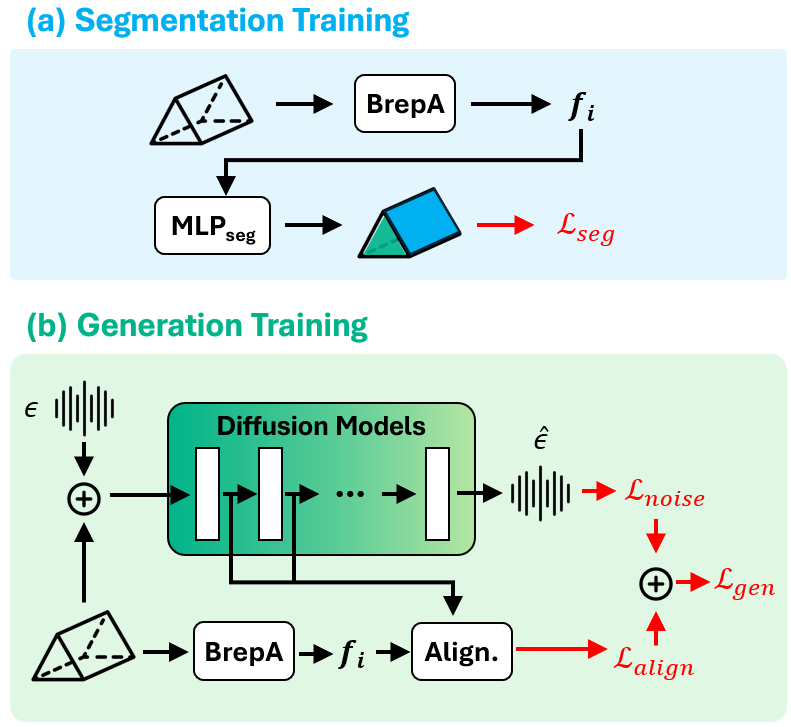}
    \caption{(a) \textbf{Segmentation Training:} The pretrained BRepA encoder is fine-tuned to provide surface representation $\mathbf{f}_{i}$ of node~$i$, which is fed into a segmentation head $\mathrm{MLP}_{\text{seg}}$ for surface-wise prediction.
    (b) \textbf{Generation Training:} BRepA guides the training of B-Rep diffusion models. The representation $\mathbf{f}_i$ from BRepA are aligned with intermediate diffusion representations through the alignment loss $\mathcal{L}_{\text{align}}$, combined with the standard noise prediction loss $\mathcal{L}_{\text{noise}}$ to form the total generation objective $\mathcal{L}_{\text{gen}}$.}
    \label{fig:down}
    \end{center}
\end{figure}

\subsection{One Pretraining, Two Domains}

Once BRepA is pretrained, the learned representations $\mathbf{f}$ can be effectively utilized across various downstream tasks. 
In this work, we demonstrate the benefits of BRepA in two representative categories: segmentation and generative modeling. 
Specifically, we evaluate how BRepA accelerates convergence and improves performance in both scenarios.

\paragraph{Segmentation Training.}
For segmentation tasks, we attach a lightweight four-layer $\mathrm{MLP}_{\text{seg}}$ as a segmentation prediction head on top of the pretrained BRepA encoder. 
The model is then fine-tuned using the available labeled samples, with the multi-class cross-entropy loss $\mathcal{L}_{seg}$ as the training objective, as depicted in Figure~\ref{fig:down}~(a).
Essentially, BRepA provides a well-initialized representation space that mitigates overfitting and stabilizes the optimization of the segmentation network. 
In our experiments, we show that this simple BRepA-based segmentation model consistently outperforms UV-Net~\cite{uvnet} with a comparable number of parameters.

\paragraph{Generative Training.}
For generative training, we employ the pretrained representation $\mathbf{f}$ from BRepA to guide the intermediate features of a B-Rep diffusion model, thereby accelerating its training process, as depicted in Figure~\ref{fig:down}~(b).
Following the design in~\cite{repa}, Representation Alignment (REPA) is performed on the early stacked transformer layers within the diffusion model to enforce representation consistency. 
Specifically, for the first $K$ transformer layers, we extract their intermediate outputs and apply average pooling to obtain the hidden feature $\mathbf{o}_{k,i}$ for surface $i$, where $k$ denotes the layer index. 
We introduce a feature alignment objective based on negative cosine similarity, which is incorporated into the overall diffusion training loss:
\begin{align}
\mathcal{L}_{gen} &= \mathcal{L}_{\text{noise}} + \beta \mathcal{L}_{\text{align}}, \label{eq:total_loss} \\[4pt]
\mathcal{L}_{\text{noise}} &= \text{MSE}(\boldsymbol{\epsilon}, \boldsymbol{\epsilon}'), \label{eq:noise_loss} \\[4pt]
\mathcal{L}_{\text{align}} &= \sum_i \sum_k \Big( 1 - \frac{\mathbf{o}_{k,i} \cdot \mathbf{f}_i}{\|\mathbf{o}_{k,i}\| \, \|\mathbf{f}_i\|} \Big), \label{eq:align_loss}
\end{align}
where $\mathcal{L}_{\text{noise}}$ is the standard noise prediction term in diffusion training, while $\mathcal{L}_{\text{align}}$ enforces alignment between the pretrained BRepA features and the diffusion model representations. The coefficient $\beta$ balances the contribution of the two loss terms. 
It is worth noting that, despite its effectiveness, BRepA (13.5M) contains around 5\% of the parameters compared to current state-of-the-art B-Rep diffusion models, BrepDiff~\citep{brepdiff} (260M), highlighting its efficiency and generalization capability.

\section{Experiments}

\newcommand{\valdelta}[2]{%
  \begin{tabular}[c]{@{}c@{}}#1\\[-2pt]{\scriptsize (#2)}\end{tabular}
}
\begin{table*}[t]
\caption{\textbf{Segmentation performance on the MFCAD~\cite{cao2020graph} dataset under different label ratios.} 
Results are reported in IoU (\%). 
The table compares UV-Net baselines with BRepA variants, including configurations without self-supervised learning (SSL), with only geometry MAE, and with full dual-MAE pretraining.
Across all label ratios, BRepA with dual-MAE pretraining consistently achieves the best performance. Notably, under the setting of 1\% labeled samples, dual-MAE pretraining yields a substantial improvement of \textbf{32.52\%} in mIoU.
}
\centering
\renewcommand{\arraystretch}{1.1}
{
\small
\begin{tabular}{lcccccccc}
\toprule
\textbf{Method} & \textbf{0.10\%} & \textbf{0.20\%} & \textbf{0.50\%} & \textbf{1\%} & \textbf{10\%} & \textbf{20\%} & \textbf{50\%} & \textbf{100\%} \\ 
\midrule
UV-Net~\cite{uvnet} & 6.73 & 9.46 & 14.08 & 19.92 & 49.87 & 68.15 & 83.94 & 90.58 \\
UV-Net++~\cite{uvnet} & 7.89 & 12.17 & 18.43 & 28.06 & 63.85 & 79.24 & 90.91 & 95.88 \\
UV-Net++~\cite{uvnet} \& SDF SSL~\cite{jones2023self} & 9.42 & 14.63 & 21.87 & 34.12 & 72.94 & 87.16 & 94.73 & 97.85 \\
\midrule
BRepA (w/o SSL) & 8.98 & 14.89 & 20.31 & 35.06 & 79.15 & 92.91 & 98.18 & 99.19 \\
BRepA (w/ only geo. MAE)
& \valdelta{14.29}{+5.31}
& \valdelta{20.22}{+5.33}
& \valdelta{36.64}{+16.33}
& \valdelta{63.10}{+28.04}
& \valdelta{88.73}{+9.58}
& \valdelta{95.79}{+2.88}
& \valdelta{98.27}{+0.09}
& \valdelta{99.14}{-0.05} \\
\rowcolor{OursRow}
\textbf{BRepA (ours)}
& \valdelta{\textbf{15.87}}{+6.89}
& \valdelta{\textbf{22.76}}{+7.87}
& \valdelta{\textbf{40.33}}{+20.02}
& \valdelta{\textbf{67.58}}{+32.52}
& \valdelta{\textbf{91.24}}{+12.09}
& \valdelta{\textbf{96.95}}{+4.04}
& \valdelta{\textbf{99.12}}{+0.94}
& \valdelta{\textbf{99.20}}{+0.01} \\
\bottomrule
\end{tabular}
}
\label{tab:seg_mfcad}
\end{table*}

In this section, we first describe the experimental setup, including the datasets, training details, and evaluation metrics.
Next, we demonstrate the impact of BRepA pretraining on segmentation training, particularly when the number of labeled samples is limited.
We then show how BRepA benefits the B-Rep generation model, BrepDiff~\cite{brepdiff}, in terms of both training speed and final performance.
Finally, we investigate the effect of BRepA's model size.

\subsection{Experimental Setup}

\paragraph{Datasets.}
We conduct experiments on two public B-Rep datasets, MFCAD~\cite{cao2020graph} and Fusion 360 Segmentation~\cite{willis2020fusion}.
MFCAD provides 16 machining-feature categories, while Fusion 360 contains 8 face-level modeling-operation categories.
Following the official trainval/test splits, we use 12367/3094 samples for MFCAD and 30459/5399 for Fusion 360.
For segmentation, we randomly sample 0.1\%, 0.2\%, 0.5\%, 1\%, 10\%, 20\%, and 50\% of the training set as labeled subsets; for generation, we use the full training set.

\paragraph{Pretraining Details.}
For the pretraining of BRepA, we train the model for 1,000 epochs using an NVIDIA A6000 GPU with 48GB VRAM. 
The checkpoint with the lowest pretraining loss $\mathcal{L}_{MAE}$ on the validation set is selected for the subsequent segmentation and generation experiments. 
Training on the MFCAD dataset converges to the optimal checkpoint in approximately 2 hours, while training on the Fusion 360 dataset takes around 6 hours. 

\paragraph{Segmentation \& Generation Training Details.}
For the segmentation task, we append a four-layer segmentation head $\mathrm{MLP}_{\text{seg}}$ on top of the pretrained BRepA encoder and fine-tune the entire model. 
We compare BRepA with the following baselines: 
(1) \textbf{UV-Net}: the official model and default configuration from~\cite{uvnet}; 
(2) \textbf{UV-Net++}: an enlarged version of UV-Net with the same parameter count as BRepA, trained for the same number of epochs as BRepA's pretraining \& finetuning; 
(3) \textbf{UV-Net++ w/ SDF SSL}: the UV-Net++ baseline further pretrained with the SDF reconstruction objective proposed in~\cite{jones2023self}; 
(4) \textbf{BRepA (w/o SSL)}: BRepA with $\mathrm{MLP}_{\text{seg}}$ but trained from scratch without any pretraining; 
(5) \textbf{BRepA (w/ only geo. MAE)}: BRepA pretrained only with the geometric MAE objective.

\vspace{0.5em}
For the generative task, we adopt {BrepDiff}~\cite{brepdiff} as the base diffusion model and use its official implementation and training configurations as our baseline. 
In addition, following the feature alignment strategy described in Section~\ref{subsec:dual_mae}, 
we align the outputs of the first four transformer layers (out of twelve) in BrepDiff~\cite{brepdiff} with the pretrained representations from BRepA.

\paragraph{Metrics.}
During the pretraining stage of BRepA, we select the model checkpoint with the lowest $\mathcal{L}_{\text{MAE}}$ for the subsequent segmentation and generation experiments. 
For the segmentation tasks, we report the results in terms of accuracy (Acc) and mean Intersection-over-Union (mIoU). 
For the generative tasks, we evaluate 1-nearest neighbor accuracy (1-NNA), coverage (COV), and minimum matching distance (MMD). 
The detailed computation formulas of these metrics are provided in the supplementary materials.

\subsection{Experimental Results}

\begin{table*}[th]
\caption{\textbf{Segmentation performance on the Fusion 360~\cite{willis2020fusion} dataset under different label ratios.}
Results are reported in mIoU (\%).
The table compares UV-Net baselines with BRepA variants, including configurations without self-supervised learning (SSL), with only geometric MAE, and with full dual-MAE pretraining.
Across all label ratios, BRepA with dual-MAE pretraining consistently achieves the best performance, and notably, under the setting of 0.5\% labeled samples, dual-MAE pretraining yields a considerable improvement of \textbf{13.63\%} in mIoU.
}
\centering
\renewcommand{\arraystretch}{1.1}
{
\small
\begin{tabular}{lcccccccc}
\toprule
\textbf{Method} & \textbf{0.10\%} & \textbf{0.20\%} & \textbf{0.50\%} & \textbf{1\%} & \textbf{10\%} & \textbf{20\%} & \textbf{50\%} & \textbf{100\%} \\ 
\midrule
UV-Net \cite{uvnet} & 4.23 & 6.15 & 7.84 & 11.92 & 36.47 & 41.83 & 53.76 & 61.34 \\
UV-Net++ \cite{uvnet} & 5.87 & 8.24 & 10.93 & 13.68 & 41.12 & 48.35 & 57.64 & 62.85 \\
UV-Net++ \cite{uvnet} \& SDF SSL \cite{jones2023self} & 8.91 & 9.76 & 14.25 & 16.83 & 43.58 & 51.47 & 58.32 & 63.15 \\
\midrule
BRepA (w/o SSL) & 9.21 & 11.90 & 15.57 & 19.11 & 48.19 & 53.22 & 61.10 & 65.20 \\
BRepA (w/ only geo. MAE)
& \valdelta{14.92}{+5.71}
& \valdelta{18.73}{+6.83}
& \valdelta{23.26}{+7.69}
& \valdelta{25.84}{+6.73}
& \valdelta{52.35}{+4.16}
& \valdelta{55.78}{+2.56}
& \valdelta{61.94}{+0.84}
& \valdelta{66.27}{+1.07} \\
\rowcolor{OursRow}
\textbf{BRepA (ours)}
& \valdelta{\textbf{17.94}}{+8.73}
& \valdelta{\textbf{24.16}}{+12.26}
& \valdelta{\textbf{29.20}}{+13.63}
& \valdelta{\textbf{32.04}}{+12.93}
& \valdelta{\textbf{55.27}}{+7.08}
& \valdelta{\textbf{58.69}}{+5.47}
& \valdelta{\textbf{63.97}}{+2.87}
& \valdelta{\textbf{67.19}}{+1.99} \\
\bottomrule
\end{tabular}
}
\label{tab:seg_fusion}
\end{table*}

\begin{figure}[t]
    \centering
    \includegraphics[width=0.85\linewidth]{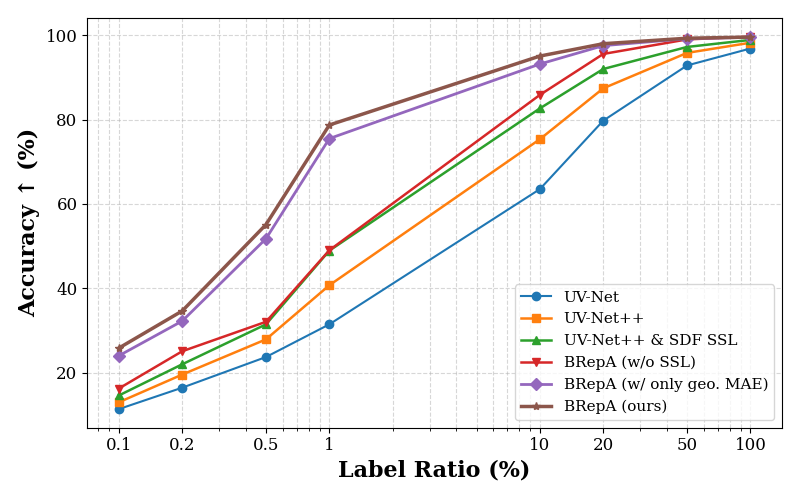}
    \caption{\textbf{Comparison of segmentation accuracy across varying label ratios on the MFCAD~\cite{cao2020graph} dataset.}
    Our BRepA consistently outperforms UV-Net and its variants, especially under low-label regimes, demonstrating strong label-efficiency.
    The combination of geometric and topology MAE pretraining yields the highest overall accuracy across all label ratios.}
    \label{fig:seg_mfcad_acc}
\end{figure}

\begin{figure}[t]
    \centering
    \includegraphics[width=0.85\linewidth]{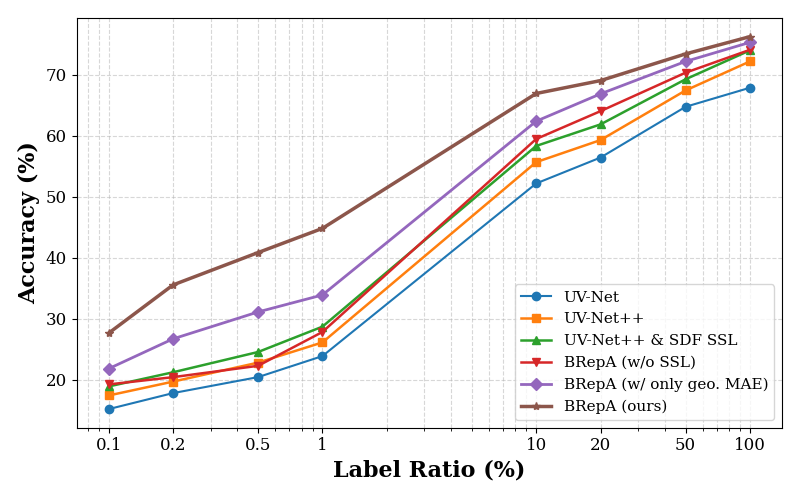}
    \caption{\textbf{Comparison of segmentation accuracy across varying label ratios on the Fusion 360 segmentation~\cite{willis2020fusion} dataset.}
    Our BRepA consistently outperforms UV-Net and its variants.
    The combination of geometric and topology MAE pretraining yields the highest overall accuracy across all label ratios.}
    \vspace{1em}
    \label{fig:seg_fusion_acc}
\end{figure}

\paragraph{Segmentation Training.}
The mIoU results of BRepA and the baseline models on MFCAD and Fusion 360 are reported in Tables~\ref{tab:seg_mfcad} and \ref{tab:seg_fusion}, respectively, while the accuracy results are presented in Figures~\ref{fig:seg_mfcad_acc} and \ref{fig:seg_fusion_acc}. Across all label ratio settings, full BRepA training consistently achieves the best performance on both mIoU and accuracy. Moreover, the benefits of BRepA pretraining are particularly pronounced when labeled data are scarce: with only 1\% of MFCAD labeled, BRepA pretraining improves mIoU by \textbf{32.52\%}, and with only 0.5\% of Fusion 360 labeled, it improves mIoU by \textbf{13.63\%}. In addition, the advantages of BRepA (w/o SSL) over UV-Net++ highlight the effectiveness of novel modules such as Transformer layers and GAT compared to using only CNNs and vanilla GCNs.
Focusing solely on geometry MAE, the performance gains brought by BRepA's geometry MAE are significantly larger than those from SDF SSL~\cite{jones2023self}; moreover, the inclusion of topology MAE in BRepA further improves the model's performance.
In Figure~\ref{fig:seg_viz}, we present several samples from the Fusion 360 segmentation dataset~\cite{willis2020fusion}, including the segmentation ground truth and predictions from (1) BRepA + MLP$_{\text{seg}}$ and (2) UV-Net++~\cite{uvnet} with SDF SSL~\cite{jones2023self} pretraining, under the setting where 100\% of the labels are used. The results demonstrate that BRepA + MLP$_{\text{seg}}$ achieves better higher prediction accuracy compared to the baseline method.
Additional qualitative results are provided in the supplementary material.

\begin{figure}[t]
    \centering
    \includegraphics[width=0.8\linewidth]{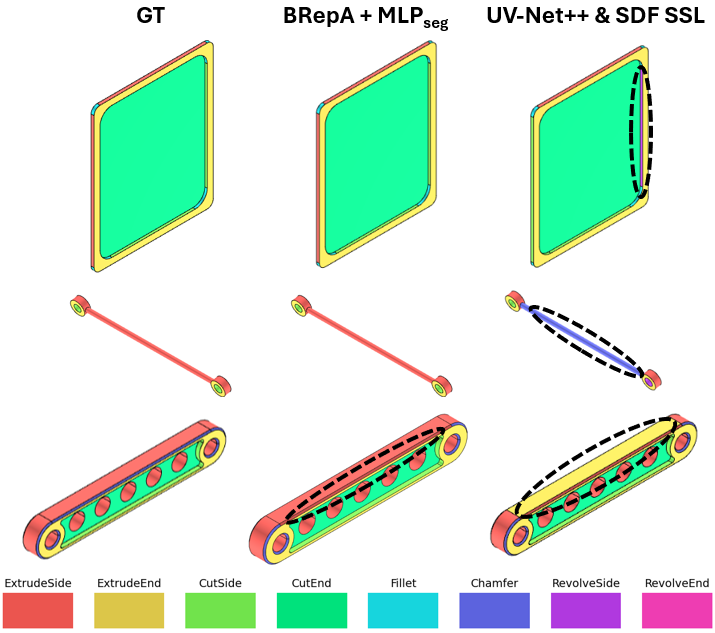}
    \caption{\textbf{Qualitative comparison of segmentation results on the Fusion 360 segmentation~\cite{willis2020fusion} dataset} (100\% of the labels are used). The visualization shows ground truth (GT) across different CAD models, along with the prediction from (1) BRepA + MLP$_{\text{seg}}$, and (2) UV-Net++~\cite{uvnet} with SDF SSL pretraining~\cite{jones2023self}. Incorrectly predicted surfaces are highlighted with black dashed circles.}
    \label{fig:seg_viz}
\end{figure}

\begin{figure}[t]
    \centering
    \includegraphics[width=0.8\linewidth]{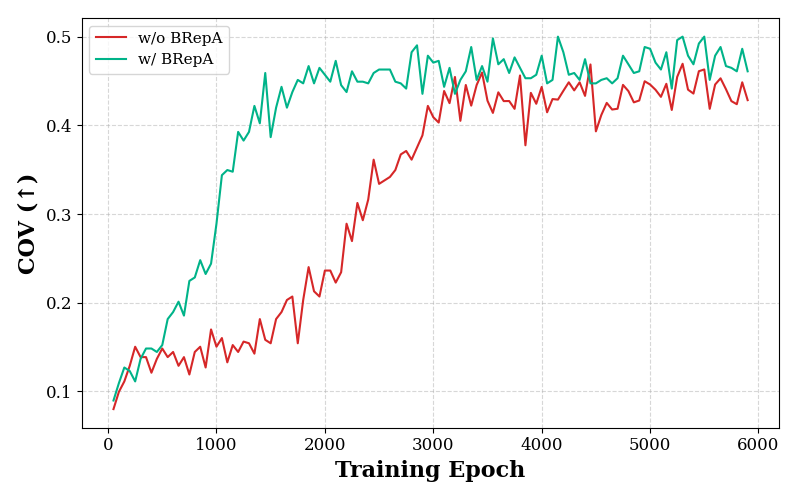}
    \caption{\textbf{Training convergence comparison of BrepDiff~\cite{brepdiff} with and without BRepA feature alignment.} The integration of BRepA accelerates training speed by nearly 2× while achieving superior final coverage (COV) performance (higher is better).}
    \label{fig:brepdiff_cov}
\end{figure}

\begin{figure}[t]
    \centering
    \includegraphics[width=0.8\linewidth]{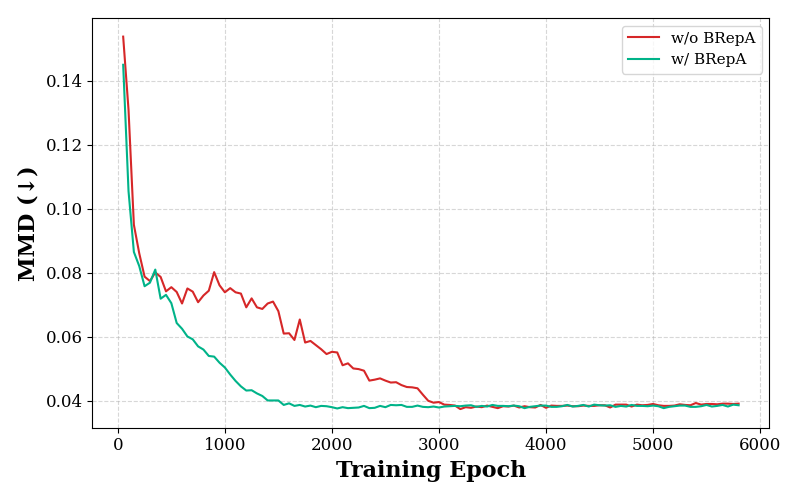}
    \caption{\textbf{Training convergence comparison of BrepDiff~\cite{brepdiff} with and without BRepA feature alignment in terms of MMD.} The integration of BRepA accelerates training speed by nearly 2× (for MMD, lower is better).}
    \label{fig:brepdiff_mmd}
\end{figure}

\paragraph{Generation Training.}
Adopting the default training method and the training augmented with BRepA-based feature alignment, the generative performance of BrepDiff~\cite{brepdiff} is reported in Table~\ref{tab:gen_mfcad}.
On both the MFCAD~\cite{cao2020graph} and Fusion~360~\cite{willis2020fusion} datasets, incorporating BRepA consistently improves the generation quality.
For example, in terms of 1-NNA, BRepA brings gains of \textbf{3.3\%} and \textbf{3.5\%} on the two datasets, respectively.
Importantly, the benefit of BRepA in accelerating training is particularly significant. As is well known, diffusion models~\cite{ho2020denoising} are computationally expensive to train; however, BRepA effectively alleviates this issue.
During training, we perform validation every 50 epochs by generating samples and computing 1-NNA, COV, and MMD. The evolution of these metrics over training epochs is shown in Figure~\ref{fig:teaser}~(b), \ref{fig:brepdiff_cov}, and \ref{fig:brepdiff_mmd}.
All three figures consistently demonstrate that integrating BRepA leads to a substantial acceleration in BrepDiff training, allowing it to reach strong performance using nearly half the training time, while also improving the final convergence quality, as reflected by better 1-NNA and COV results.
It is worth noting that the default BRepA model contains only 13.5M parameters, whereas BrepDiff has roughly 260M parameters, almost \textbf{20x} larger.
The fact that such a compact model can substantially accelerate a generative model far larger than itself highlights the strong generalization of the representations learned by BRepA pretraining.

\begin{table}[t]
\caption{\textbf{Generative performance comparison of BrepDiff~\cite{brepdiff} and on the MFCAD~\cite{cao2020graph} and Fusion 360~\cite{willis2020fusion} datasets}, 
trained with and without BRepA representations. 
Results are reported in terms of 1-NNA, Coverage (COV), and Minimum Matching Distance (MMD).
The symbol $^{\dagger}$ indicates that BrepDiff is trained with BRepA feature alignment.}
\centering
\resizebox{\linewidth}{!}{
\renewcommand{\arraystretch}{1.2}
\begin{tabular}{lcccc}
\toprule
\textbf{Dataset} & \textbf{Model} & \textbf{1-NNA} ($\downarrow$) & \textbf{COV} ($\uparrow$) & \textbf{MMD} ($\downarrow$) \\ 
\hline
\multirow{2}{*}{MFCAD} 
& BrepDiff & 0.574 & 0.469 & 0.038 \\ 
& BrepDiff $^{\dagger}$ & 0.541 & 0.500 & 0.037 \\[2pt] \hline
\multirow{2}{*}{Fusion 360} 
& BrepDiff & 0.577 & 0.486 & 0.026 \\ 
& BrepDiff $^{\dagger}$ & 0.542 & 0.510 & 0.025 \\
\bottomrule
\end{tabular}
}
\label{tab:gen_mfcad}
\end{table}

\begin{table}[t]
\centering
\caption{\textbf{Effect of model scaling on BRepA performance on the MFCAD~\cite{cao2020graph} dataset.}
Larger variants (BRepA-L and BRepA-XL) are obtained by uniformly increasing the feature dimensions and attention heads in all encoders. 
Scaling up BRepA consistently improves mIoU across all label ratios, demonstrating the strong scalability of BRepA.}
\vspace{3pt}
\small
\renewcommand{\arraystretch}{1.1}{
\begin{tabular}{lcccc}
\toprule
\textbf{Model} & \textbf{Params} & \textbf{0.50\%} & \textbf{1\%} & \textbf{10\%} \\
\midrule
{BRepA}     & 13.5M & 40.33 & 67.58 & 91.24 \\
{BRepA-L}   & 27M   & 42.81 & 69.63 & 92.70 \\
{BRepA-XL}  & 80M   & 44.37 & 71.58 & 93.97 \\
\bottomrule
\end{tabular}
}
\label{tab:scale}
\end{table}

\subsection{Experimental Analysis}

For deep learning models, larger parameter sizes often correlate with stronger capacity and improved performance.
Under the default configuration, BRepA contains only 13.5M parameters. To investigate the effect of model scaling, we increase the dimensionality of intermediate features, the number of attention heads, and the Transformer layers to construct two larger variants: BRepA-L~\textbf{(27M)} and BRepA-XL~\textbf{(80M)}.
Importantly, we scale the surface, edge, and graph encoders uniformly; increasing the capacity of only a single module leads to unstable training. We evaluate both BRepA-L and BRepA-XL on the MFCAD dataset, and the results are reported in Table~\ref{tab:scale}.
As observed from Table~\ref{tab:scale}, scaling up BRepA consistently improves mIoU across all label ratios. In particular, the largest variant, BRepA-XL, achieves the highest mIoU values for all label ratios, indicating that increasing model capacity allows for better feature representation and stronger generalization.

\section{Conclusion}

In this work, we introduced BRepA, a unified self-supervised pretraining framework that addresses B-Rep CAD learning challenges through dual masked autoencoding of geometric and topological information. Our experiments demonstrate substantial improvements across both discriminative and generative tasks, achieving 32.52\% mIoU improvement with only 1\% labeled samples and accelerating diffusion model training by nearly 2× despite being 20× smaller. The success stems from our novel architecture combining Vision Transformers for geometric features with Graph Attention Networks for topological modeling, enabling effective cross-domain generalization. Overall, BRepA establishes a powerful foundation for CAD AI systems, demonstrating that geometry- and topology-aware pretraining provides a versatile backbone for both efficient generative model training and improved discriminative performance with limited labels.

\maketitlesupplementary
\setcounter{section}{0}
\renewcommand{\thesection}{\Alph{section}}
\tableofcontents

\section{Evaluation Metric}

Given a generated set $\mathcal{G} = \{ x^g \}$ and a real dataset $\mathcal{R} = \{ x^r \}$, where $x^g$ and $x^r$ denote generated and real samples respectively, we evaluate the generative model by comparing these two sets. In practice, $\mathcal{R}$ corresponds to the test set that is not used during generative model training, while $\mathcal{G}$ is the collection of samples produced at inference time. The distance between two samples is measured using the Chamfer distance $D(\cdot)$. Based on this distance, the formulas for metrics including coverage (COV), minimum matching distance (MMD)~\citep{achlioptas2018learning}, and 1-nearest neighbor accuracy (1-NNA)~\citep{yang2019pointflow} are defined as follows:

\paragraph{Coverage (COV)} quantifies how well the generated set $\mathcal{G}$ covers the real dataset $\mathcal{R}$. For each $x^g \in \mathcal{G}$, it selects its nearest neighbor in $\mathcal{R}$, forming a subset of real samples. COV is then defined as the ratio between the size of this subset and the cardinality of $\mathcal{R}$:
\begin{equation}
    \text{COV}(\mathcal{G}, \mathcal{R}) = \frac{| \{ \arg\min_{x_r \in \mathcal{R}} D(x_g, x_r ) | x_g \in \mathcal{G} \} |}{|\mathcal{R}|}.
\end{equation}

\paragraph{Minimum Matching Distance (MMD)} measures the fidelity of the generated set by computing, for each $x^r \in \mathcal{R}$, the distance to its nearest neighbor in $\mathcal{G}$. MMD is defined as the average of these distances:
\begin{equation}
    \text{MMD}(\mathcal{G}, \mathcal{R}) = \frac{1}{|\mathcal{R}|} \sum_{x_r \in \mathcal{R}} \min_{x_g \in \mathcal{G}} D(x_g, x_r).
\end{equation}

\paragraph{1-Nearest Neighbor Accuracy (1-NNA)} assesses the similarity between $\mathcal{R}$ and $\mathcal{G}$ by computing the proportion of samples in either set whose nearest neighbor comes from the same set. A lower 1-NNA indicates greater distributional overlap between real and generated samples:
\small
\begin{equation}
    \text{1-NNA}(\mathcal{G}, \mathcal{R}) = \frac{\sum_{x_g \in \mathcal{G}} \mathbbm{1}(N_{x_g} \in \mathcal{G}) + \sum_{x_r \in \mathcal{R}} \mathbbm{1}(N_{x_r} \in \mathcal{R})}{|\mathcal{G}|+|\mathcal{R}|},
    \label{eq:1-nna-supp}
\end{equation}
\normalsize
where $\mathbbm{1}[\cdot]$ denotes the indicator function, and $N_{x_g}$ is the nearest neighbor of $x_g$ within the set $\mathcal{R} \cup \mathcal{G} \setminus {x_g}$; the same definition applies to $N_{x_r}$. When $\mathcal{G}$ closely matches $\mathcal{R}$, it becomes difficult to distinguish whether the nearest neighbor of a sample $x_g$ comes from $\mathcal{G}$ or $\mathcal{R}$, and vice versa. In this scenario, the 1-NNA score converges toward \textbf{50\%}.
\section{Additional Experimental Details and Results}

\subsection{Few-Shot Classification with BRepA}

In the main text, we present the utility of BRepA for discriminative tasks by demonstrating its substantial impact on segmentation performance. To further illustrate its broad applicability, we evaluate BRepA on another discriminative task: classification. We conduct experiments on the SolidLetters~\cite{uvnet} dataset under highly challenging 1-, 2-, and 5-shot settings. The experimental results, as summarized in Table~\ref{tab:classification}, further confirm the effectiveness of BRepA across diverse discriminative scenarios.

\begin{table}
\centering
\caption{\textbf{Few-shot classification accuracy (\%) on the SolidLetters dataset.} BRepA consistently enhances performance across all settings, demonstrating its strong generalization as a discriminative feature extractor.}
\label{tab:classification}
\begin{tabular}{l c c c }
\toprule
 & 1-shot~~ & 2-shot~~ & 5-shot \\
\hline \hline
w/ BRepA & 12.40 & 27.13 & 58.95 \\
w/o BRepA & 11.83 & 17.76 & 38.37 \\
\bottomrule
\end{tabular}
\end{table}

\subsection{Robustness of BRepA across Different Generative Architectures and Datasets}
\begin{figure}[t]
    \centering
    \includegraphics[width=\linewidth]{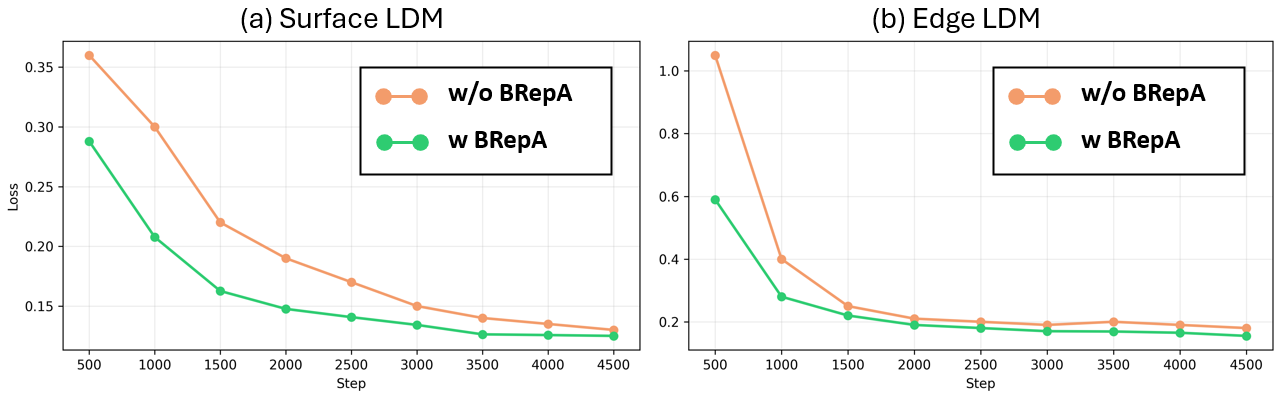}
    \caption{\textbf{Loss curves for the Surface LDM (a) and Edge LDM (b) during the training of BrepGen.} The results demonstrate that integrating BRepA leads to significantly faster convergence across both generative components.}
    \label{fig:noise_curves}
\end{figure}

In the main text, we demonstrated that BRepA significantly accelerates the training of the B-Rep generative model BrepDiff~\cite{brepdiff}. Here, we further show that BRepA similarly enhances the training efficiency of another B-Rep generative model, BrepGen~\cite{brepgen}, thereby highlighting the robust generalization capability of BRepA in accelerating the training of generative models. BrepGen employs separate Latent Diffusion Models (LDMs) for surfaces and edges; accordingly, we align BRepA features with each of these components. The validation noise curves for these two LDMs during training are illustrated in Fig.~\ref{fig:noise_curves}, indicating that both LDMs achieve noticeably faster convergence.

\begin{figure}
    \centering
    \includegraphics[width=0.5\linewidth]{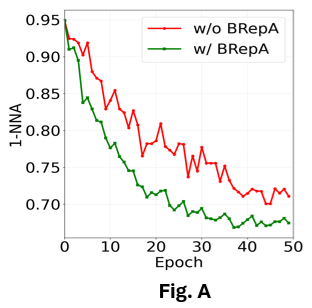}
    \caption{\textbf{Training convergence of BrepDiff on ABC.} Pretrained BRepA accelerates convergence and improves the 1-NNA metric, demonstrating its generalization to diverse B-Rep distributions.}
    \label{fig:brepdiff_abc}
\end{figure}

Additionally, we select MFCAD and Fusion 360 in the manuscript because they provide native segmentation labels, allowing us to demonstrate that pretrained BRepA can be effectively applied to both understanding and generation tasks. To address generalization to more complex B-Rep distributions, we additionally evaluate BRepA on the larger and more diverse ABC~\cite{koch2019abc} dataset. As shown in Figure~\ref{fig:brepdiff_abc}, BRepA substantially accelerates the training of BrepDiff on ABC.

\subsection{Effect of Mask Ratio}
In the dual-MAE pretraining of BRepA, we randomly mask either surface nodes or edges in the B-Rep graph.
In this experiment, we investigate how different masking strengths affect the pretraining effectiveness of BRepA. 

\begin{figure}[t]
    \centering
    \includegraphics[width=\linewidth]{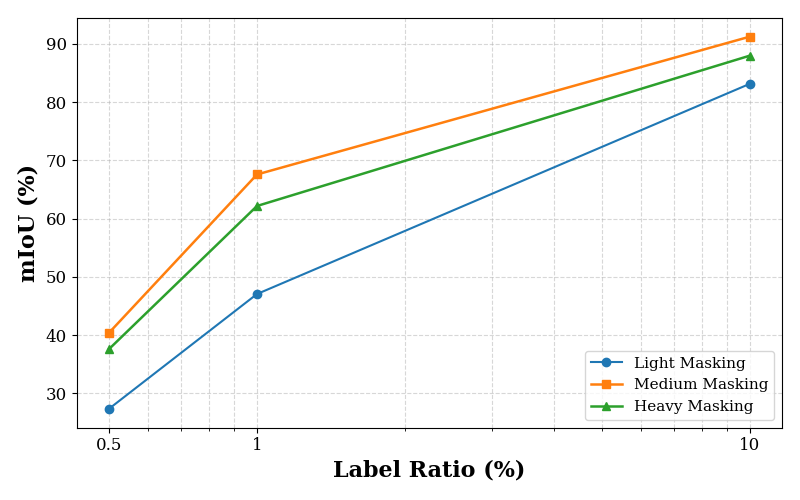}
    \caption{\textbf{Comparison of segmentation performance (mIoU) under different masking strengths in BRepA’s dual-MAE pretraining.}
    Medium masking consistently achieves the best performance across different label ratios (0.5\%, 1\%, 10\%).}
    \label{fig:mask_ratio}
\end{figure}

For \textbf{light} masking, at each iteration we sample a mask ratio $m\%$ from a uniform distribution $[0, 0.2]$ and randomly mask $m\%$ of the surface nodes and edges.
Similarly, for \textbf{medium} and \textbf{heavy} masking, we sample $m\%$ from $[0, 0.5]$ and $[0, 0.8]$, respectively. 
We conduct repeated experiments on the MFCAD~\cite{cao2020graph} dataset under the \{0.5\%, 1\%, 10\%\} label-ratio settings. The results are reported in Fig.~\ref{fig:mask_ratio}.

Medium masking consistently delivers the best performance across all label-ratio settings, indicating that a moderate level of corruption provides the strongest pretraining signal. Light masking fails to sufficiently challenge the model, resulting in weaker representation learning and notably lower mIoU. In contrast, heavy masking overly disrupts geometric and topological structure, causing a drop in performance compared to the medium-masking regime.
Therefore, in all subsequent experiments, we uniformly sample the mask ratio from the interval $[0, 0.5]$.

\subsection{Robustness of BRepA in Segmentation Training}
For the segmentation experiments in the manuscript, we randomly selected labeled subsets comprising 0.10\%, 0.20\%, 0.50\%, 1\%, 10\%, 20\%, and 50\% of the dataset.
In these experiments, BRepA with full dual MAE pretraining consistently achieved the best performance across all label ratios. To verify that these results were not influenced by the randomness of the labeled subset selection, we repeated the experiment with a 1\% label ratio five times. The results are summarized in Table~\ref{tab:supp_repeat}.
The experimental results show that BRepA with dual MAE pretraining consistently achieves the best performance.
Compared to the 2D SDF SSL method~\cite{jones2023self}, it exhibits slightly lower standard deviation, indicating greater stability.

\begin{table}[t]
\caption{\textbf{Repeated segmentation results of BRepA and baseline models on a 1\% labeled subset.} 
Each model was evaluated five times to assess robustness to the randomness of subset selection. 
The table reports the mean and standard deviation of mIoU (\%).}
\centering
\renewcommand{\arraystretch}{1.12}{
\begin{tabular}{lc}
\toprule
\textbf{Method} & \textbf{mIoU (\%)} \\
\midrule
UV-Net~\cite{uvnet} & 19.85 ± 0.41 \\
UV-Net++~\cite{uvnet} & 28.12 ± 0.58 \\
UV-Net++~\cite{uvnet} \& SDF SSL~\cite{jones2023self} & 34.05 ± 0.87 \\
BRepA (w/o SSL) & 35.18 ± 0.62 \\
BRepA (w/ only geo. MAE) & 63.25 ± 0.68 \\
BRepA & 67.42 ± 0.71 \\
\bottomrule
\end{tabular}
}
\label{tab:supp_repeat}
\end{table}

\subsection{Effect of the Number of Graph Layers in BRepA}
\begin{figure}
    \centering
    \includegraphics[width=\linewidth]{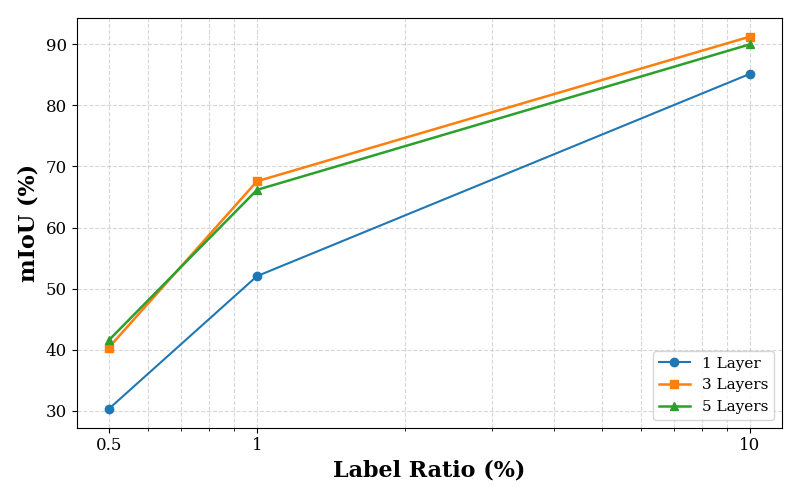}
    \caption{\textbf{Effect of the number of graph layers on segmentation performance of BRepA.} 
    The plot shows mIoU (\%) on the MFCAD dataset for different numbers of layers in the graph encoder, evaluated under 0.5\%, 1\%, and 10\% labeled subsets.
}
    \label{fig:supp_graph_layers}
\end{figure}

In the graph encoder of BRepA, we used 3 layers for both the GCN and GAT modules.
We additionally evaluate the model on the MFCAD~\cite{cao2020graph} dataset with 1, 5, and 7 layers, keeping the total number of parameters approximately the same.
When using 7 layers, BRepA consistently diverges during the early stage of training.
Therefore, in Figure~\ref{fig:supp_graph_layers}, we only report the segmentation performance for settings with 1, 3, and 5 layers on the MFCAD dataset.
The experimental results show that using multiple graph layers (e.g., 3 or 5) improves the performance of BRepA compared to a single layer, but an excessive number of layers can lead to training instability.

\subsection{Loss Balancing Weights}
In Eq.11 of the manuscript, the parameter $\alpha$ is used to balance the geometry MAE and topology MAE.
In practice, we perform variance-based rescaling to ensure that the two MAE losses are of comparable magnitude. 

For the generative training, since the primary objective of the diffusion model is noise prediction, we adjust the parameter $\beta$ in Eq.12 such that the feature alignment loss is approximately 10\% of the noise prediction loss. We additionally experiment with increasing the feature alignment loss to the same magnitude as the noise prediction loss.
This only slightly accelerates the convergence of BrepDiff, but the final generative performance actually degraded, falling below that of the BrepDiff model trained without feature alignment.

\subsection{Qualitative Results}
In Figure 7 in the manuscript, we present several samples from the Fusion 360 segmentation dataset~\cite{willis2020fusion}, including the segmentation ground truth and predictions from (1) BRepA + MLP$_{\text{seg}}$ and (2) UV-Net++~\cite{uvnet} with SDF SSL~\cite{jones2023self} pretraining, under the setting where 100\% of the labels are used. 
Here we provide more results in Figure~\ref{fig:seg_viz}.
The results demonstrate that BRepA + MLP$_{\text{seg}}$ achieves better higher prediction accuracy compared to the baseline method.

\begin{figure}
    \centering
    \includegraphics[width=0.9\linewidth]{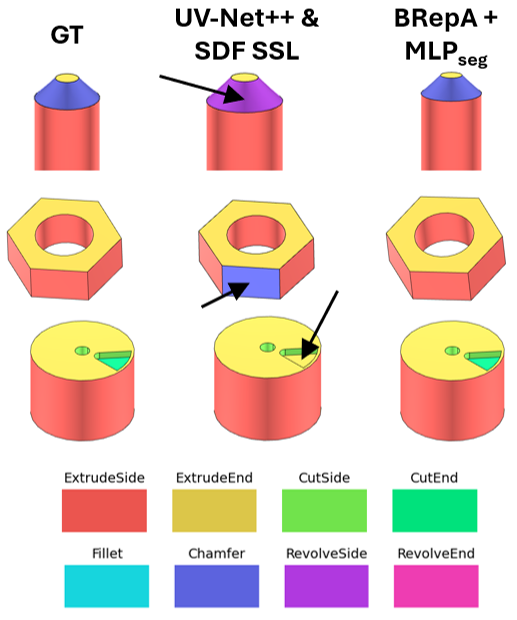}
    \caption{\textbf{Qualitative comparison of segmentation results on the Fusion 360 segmentation~\cite{willis2020fusion} dataset} (100\% of the labels are used). The visualization shows ground truth (GT) across different CAD models, along with the prediction from (1) BRepA + MLP$_{\text{seg}}$, and (2) UV-Net++~\cite{uvnet} with SDF SSL pretraining~\cite{jones2023self}. Incorrectly predicted surfaces are highlighted with black dashed circles.}
    \label{fig:seg_viz}
\end{figure}

{
    \small
    \bibliographystyle{ieeenat_fullname}
    \bibliography{main}
}

\end{document}